\documentclass[sigconf]{acmart}

\usepackage{wrapfig,lipsum,booktabs}
\usepackage{mathrsfs}  
\usepackage{microtype}
\usepackage{graphicx}
\usepackage{subfigure}
\usepackage{booktabs} 
\usepackage{verbatim}
\usepackage{amsmath}
\usepackage{amsfonts}
\usepackage{amsthm}

\usepackage{amssymb}
\usepackage{multirow}
\usepackage{booktabs}
\usepackage{makecell}
\usepackage{colortbl}
\usepackage{xcolor}
\usepackage{caption}
\usepackage[flushleft]{threeparttable}
\usepackage{thmtools}
\usepackage{thm-restate}

\usepackage{colortbl}
\usepackage{cellspace}

\usepackage{booktabs} 
\usepackage{multirow} 
\usepackage{siunitx} 

\usepackage{hyperref}
\usepackage{cleveref}
\usepackage{algorithm}
\usepackage{algorithmic}

\definecolor{highlightcolor}{RGB}{255, 255, 179}
\newcolumntype{H}[1]{>{\columncolor{highlightcolor}}#1}

\usepackage{xcolor}

\newcommand{\x}{\mathbf{x}}
\newcommand{\y}{\mathbf{y}}

\usepackage{amsmath,amsfonts,bm}

\def\eqref#1{equation~\ref{#1}}

\def\1{\bm{1}}

\DeclareMathAlphabet{\mathsfit}{\encodingdefault}{\sfdefault}{m}{sl}
\SetMathAlphabet{\mathsfit}{bold}{\encodingdefault}{\sfdefault}{bx}{n}

\usepackage{hyperref}
\usepackage{url}
\definecolor{cvprblue}{rgb}{0.21,0.49,0.74}
\acmConference[Conference acronym 'XX]{Make sure to enter the correct
  conference title from your rights confirmation emai}{June 03--05,
  2018}{Woodstock, NY}
\acmISBN{978-1-4503-XXXX-X/18/06}

\title{DeepMTL2R: A Library for Deep Multi-task Learning to Rank}

\begin{document}
\author{Chaosheng Dong}
\email{chaosd@amazon.com}
\affiliation{%
  \institution{Amazon}
  \country{USA}
}
\authornote{Corresponding author.}

\author{Peiyao Xiao}
\email{peiyaoxi@buffalo.edu	}
\affiliation{%
  \institution{University at Buffalo}
  \country{USA}
}

\author{Yijia Wang}
\email{yijiawan@amazon.com}
\affiliation{%
  \institution{Amazon}
  \country{USA}
}

\author{Kaiyi Ji}
\email{kaiyiji@buffalo.edu}
\affiliation{%
  \institution{University at Buffalo}
  \country{USA}
}







\begin{abstract}

This paper presents DeepMTL2R, an open-source deep learning framework for Multi-task Learning to Rank (MTL2R), where multiple relevance criteria must be optimized simultaneously. DeepMTL2R integrates heterogeneous relevance signals into a unified, context-aware model by leveraging the self-attention mechanism of transformer architectures, enabling effective learning across diverse and potentially conflicting objectives. The framework includes 21 state-of-the-art multi-task learning algorithms and supports multi-objective optimization to identify Pareto-optimal ranking models. By capturing complex dependencies and long-range interactions among items and labels, DeepMTL2R provides a scalable and expressive solution for modern ranking systems and facilitates controlled comparisons across MTL strategies. We demonstrate its effectiveness on a publicly available dataset, report competitive performance, and visualize the resulting trade-offs among objectives. DeepMTL2R is available at \href{https://github.com/amazon-science/DeepMTL2R}{https://github.com/amazon-science/DeepMTL2R}.

\end{abstract}

\maketitle

\section{Introduction} \label{sec: intro}

Multi-task Learning to Rank (MTL2R) extends traditional Learning to Rank (LTR) to settings where relevance is inherently multifaceted and may involve conflicting criteria \citep{har2002, svore2011learning, zhang2014}. While conventional LTR methods typically optimize a single objective, real-world ranking systems often rely on multiple noisy, subjective, and heterogeneous signals (e.g., clicks, streams, sales, dwell time, and user ratings). MTL2R addresses this gap by modeling several relevance criteria jointly, enabling ranking decisions that better reflect nuanced user preferences.


A variety of approaches have been proposed to tackle the complexity of multi-task ranking data. Early work focused on binary relevance modeling \citep{Elisseeff2001KernelMF, ferrandin2023multi} and pairwise label interactions \citep{gh2005}. More recent methods incorporate MTL techniques into tree-based models \citep{mahapatra2022multi,dong2025sigirmo} to better characterize trade-offs among competing criteria; in many cases, each criterion is treated as a separate objective and the goal is to identify Pareto-optimal solutions. In parallel, advances in deep learning have spurred neural ranking models that improve feature transformation and generalization \citep{dai2019, dai2018, guo2016, wu2018}. However, conventional feed-forward architectures often struggle with long-tailed distributions and do not reliably capture higher-order feature interactions \citep{qin2021neural, wang2017deep, Beutel2018LatentCM}. Transformer-based models offer a compelling alternative: their self-attention mechanism can represent long-range dependencies and list-level context more effectively. Despite this progress, existing methods often struggle to (i) capture context dependencies within ranked lists, (ii) scale to large datasets, and (iii) model complex relationships among labels.

\begin{figure*}
    \centering
    \includegraphics[width=0.7\textwidth]{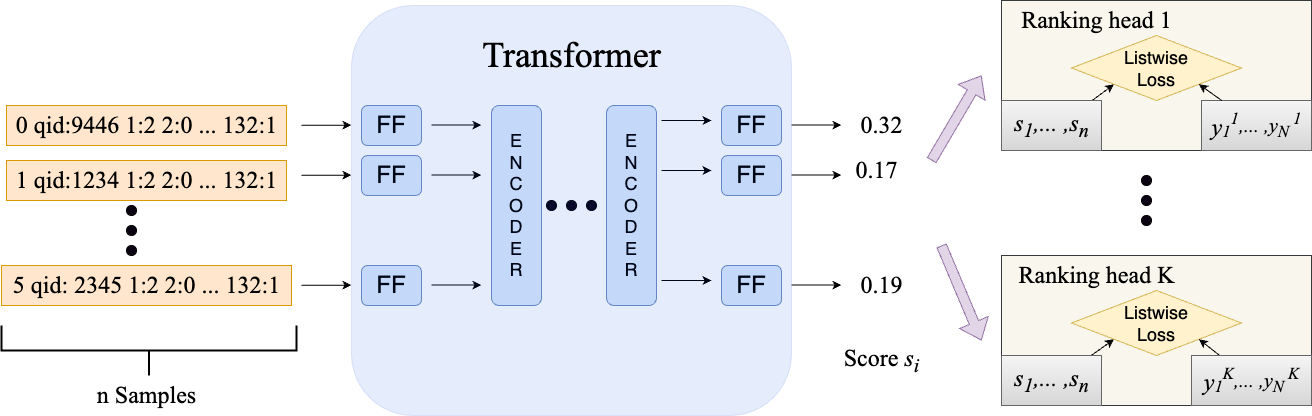} 
    \caption{DeepMTL2R: Transformer-Based DNN in MTL2R}
    \label{fig:myimage}
\end{figure*}

To address these challenges, we propose DeepMTL2R, a transformer-based neural framework for MTL2R. DeepMTL2R leverages self-attention to explicitly model list-level context and long-range dependencies among items while jointly learning task-specific relevance signals. In addition, the framework integrates multi-objective MTL techniques to learn Pareto-optimal ranking models, enabling practitioners to select solutions that best match downstream constraints and product goals. Beyond the model architecture, DeepMTL2R is designed as a research and benchmarking toolkit: it provides modular components for data preprocessing, training, evaluation, and visualization, making it straightforward to compare methods under consistent experimental settings.
We summarize our main contributions as follows:
\begin{itemize}
\item \textbf{An open-source neural framework for MTL2R:} DeepMTL2R provides a context-aware architecture that models item relevance across multiple tasks using transformers, enabling shared representation learning with a unified scoring model across tasks. The framework supports listwise training with task-specific objectives (i.e., different loss functions) and is designed to scale to large ranking datasets through efficient batching and modular components.
\item \textbf{A systematic evaluation of MTL techniques for neural MTL2R:} We implement and benchmark widely used MTL methods (e.g., Linear Scalarization and MGDA), and provide an extensible interface for adding new optimization strategies. Using the WEB30K dataset, we analyze how different methods trade off multiple relevance signals, quantify performance across tasks, and characterize the resulting Pareto front of ranking models.
\end{itemize}

\section{Problem Formulation}

\subsection{Learning to Rank}

Let $X$ denote the training set, consisting of pairs $(\mathbf{x}, \mathbf{y})$, where $\mathbf{x}$ is a list of feature vectors $\x_i \in \mathbb{R}^m$ and $\mathbf{y}$ is the corresponding list of relevance labels $y_i$, for $ i = 1, \ldots, N $. We note that $\mathbf{x}$ in the training set may have variable lengths. Let $\bm\theta$ denote the model parameters.

The goal of the learning to rank problem is to find a scoring function $f$ that optimizes a chosen Information Retrieval (IR) metric, such as Normalized Discounted Cumulative Gain (NDCG), on the test set. The scoring function $f$ is trained to minimize the mean of a surrogate loss $l$ across the training data:
\[
\mathcal{L}_{single}(\bm{\theta}) = \frac{1}{|X|} \sum_{(\x, \y) \in X} l( f_{\bm{\theta}}(\x), \y).
\]

\subsection{Multi-task Learning to Rank}

In MTL2R, multiple relevance criteria are measured for each item, yielding multiple labels for each feature vector $\x_i \in \mathbb{R}^m$. As in standard LTR, the goal is to learn a scoring function $f(\x)$ that assigns a scalar score to each item $\x_i$.
We consider a set of training examples $\x_i \in \mathbb{R}^m$, where $ i = 1, \ldots, N$. Each $\x_i$ is associated with a vector of relevance labels:
\[
\y_i = \left(y_i^1, \ldots, y_i^K\right),
\]
where $K$ denotes the number of labels (tasks). In MTL2R, the objective is to minimize a vector-valued loss:
$$
\mathcal{L}(\bm{\theta}) = [\mathcal{L}_1(\bm{\theta}), \mathcal{L}_2(\bm{\theta}), \ldots, \mathcal{L}_K(\bm{\theta})],
$$
where
$
\mathcal{L}_k(\bm{\theta}): \mathbb{R}^m \rightarrow \mathbb{R}, \forall k \in [K],
$
each tailored to a specific label.

For a minimization MTL problem, a solution $\bm{\theta}$ is referred to as \emph{non-dominated} or \emph{Pareto-optimal} if there exists no alternative solution $\bm{\theta}'$ such that $\mathcal{L}(\bm{\theta}') \preceq \mathcal{L}(\bm{\theta})$ and $\mathcal{L}(\bm{\theta}') \neq \mathcal{L}(\bm{\theta})$. The set of all Pareto-optimal solutions is called the \emph{Pareto set}. Its image under $\mathcal{L}$ is known as the \emph{Pareto front (PF)}.

\section{DeepMTL2R Framework}

\subsection{Self-Attention Mechanism}


We adopt the self-attention variant known as Scaled Dot-Product Attention\cite{NIPS2017_3f5ee243}. The attention weights are computed by taking the dot product of the query and key matrices and scaling by the square root of the model dimension. Let $\bm{Q}\in\mathbb{R}^{l\times d_{\text{model}}}$ denote the query matrix for a list of length $l$, and let $\bm{K}$ and $\bm{V}$ denote the corresponding key and value matrices. The attention output is
$$
\operatorname{Attention}(\bm{Q}, \bm{K}, \bm{V})=\operatorname{softmax}\left(\frac{\bm{Q}\bm{K}^T}{\sqrt{d_{\text {model }}}}\right)\bm{V}.
$$

\begin{figure*}
    \centering
    \includegraphics[width=0.9\linewidth]{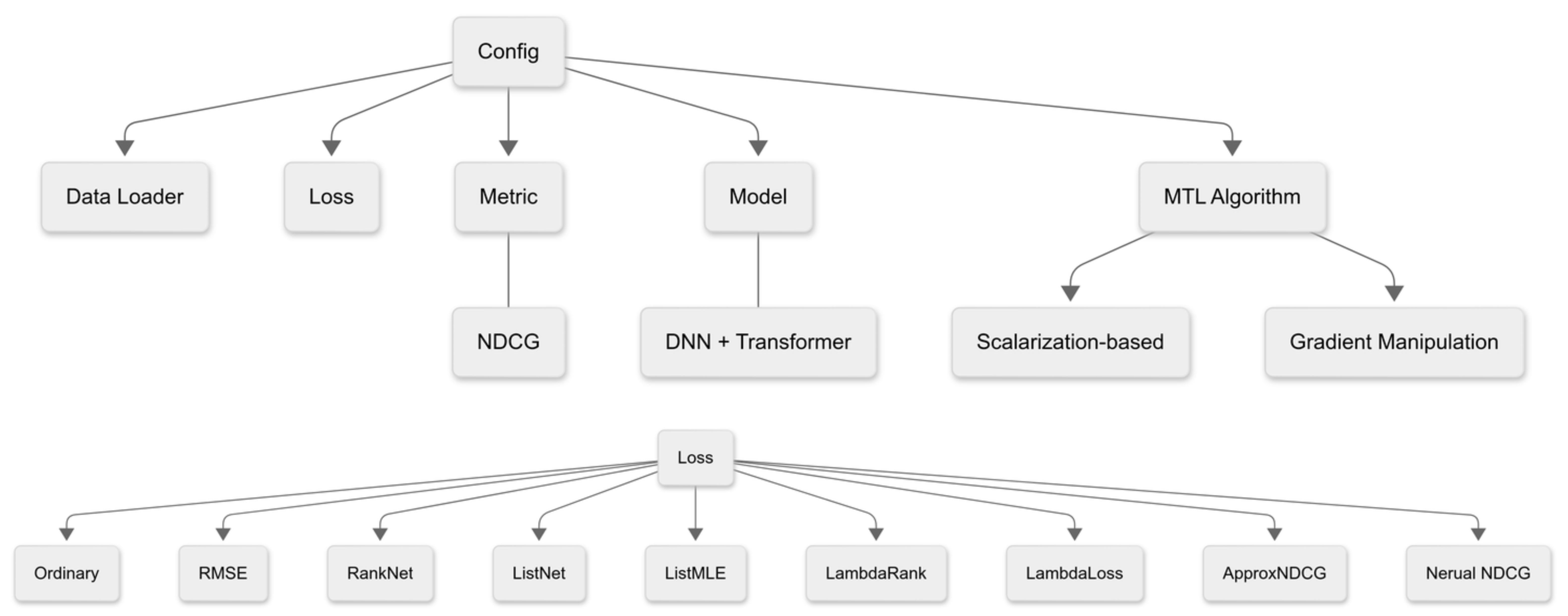}
    \caption{Structure of the DeepMTL2R library. We support 9 popular LTR losses and 21 MTL algorithms.}
    \label{fig:package}
\end{figure*}

\subsection{Model Architecture}
Items in a list are treated as tokens, and item features are viewed as token embeddings. Given an input list of length \(l\) with feature dimension \(d_f\), each item is first projected through a shared fully connected (FC) layer to dimension \(d_{fc}\). The resulting representations are then passed through a Transformer encoder comprising \(N\) encoder blocks. Each block uses multi-head self-attention with \(H\) heads and hidden dimension \(d_h\), followed by a position-wise feed-forward layer; both sublayers are wrapped with residual connections, dropout, and layer normalization. Formally,
\begin{equation}
f_{\bm{\theta}}(\x)=\mathrm{FC}(\underbrace{\operatorname{Encoder}(\operatorname{Encoder}(\ldots(\operatorname{Encoder}}_{N \text { times }}(\mathrm{FC}(\x)))))),
\end{equation}
where the Encoder function is defined as:
\begin{align}
\text { Encoder }(\x) & =\operatorname{LayerNorm}(\mathbf{z}+\operatorname{Dropout}(\mathrm{FC}(\mathbf{z}))), \notag\\
\mathbf{z} & =\operatorname{LayerNorm}(\x+\operatorname{Dropout}(\operatorname{MultiHead}(\x))).
\end{align}

After \(N\) encoder blocks, a shared fully connected layer produces a scalar score for each item, representing its relevance under the corresponding query or ranking context.
An overview of the transformer-based neural MTL2R architecture is shown in Figure \ref{fig:myimage}.

\subsection{Supported MTL Algorithms}


This section summarizes the supported MTL algorithms, which we group into scalarization-based and gradient-manipulation methods. Scalarization methods reduce the vector-valued objectives to a single surrogate objective, typically via a weighted combination, whereas gradient-manipulation methods modify or project task gradients to mitigate conflicts and improve training stability. In both cases, the update direction $d\in\mathbb{R}^m$ is typically formed by combining the $K$ task gradients as
\begin{align}
d=\sum_{k=1}^Kw_k\nabla \mathcal{L}_k(\bm{\theta})=\bm{Gw},\quad \text { s.t. } \sum_{k=1}^Kw_k=1,\quad \bm{w}\in\mathbb{R}^K_+,
\end{align}
where $w_k$ is the weight for task $k$ and
$$\bm{G}=[\nabla\mathcal{L}_1(\bm{\theta}),\nabla\mathcal{L}_2(\bm{\theta}),\ldots,\nabla\mathcal{L}_K(\bm{\theta})].$$
We implement 21 widely used MTL algorithms (e.g., MGDA \cite{desideri2012multiple}) to compute $\bm{w}$ or a modified update direction $d$. Some methods explicitly target Pareto-front discovery (PF finding), while others are designed for improving multi-task convergence and performance without explicitly constructing the PF. The full list is provided in Table \ref{tab:list of MTL}, and Figure \ref{fig:package} illustrates the overall library structure.

\begin{table*}
\centering
\caption{List of supported MTL algorithms in DeepMTL2R.}
\label{tab:list of MTL}
\resizebox{0.85\textwidth}{!}{%
\begin{tabular}{@{}lcccl@{}}
\toprule
MTL Method                        & PF Finding & Scalarization or Gradient & Suggested Evaluation metric & Reference \\ \midrule
Linear Scalarization (LS)              & Y          & Scalarization             & HVI \cite{auger2012hypervolume}               & -          \\
ScaleInvariantLinearScalarization & Y          & Scalarization             & HVI               & \cite{navon2022multi}          \\
WeightedChebyshev (WC)                 & Y          & Scalarization             & HVI               & \cite{hotegni2024multi}         \\ 
Soft WeightedChebyshev            & Y          & Scalarization             & HVI               & \cite{lin2024smooth}         \\ \midrule
EPO                               & Y          & Gradient                  & HVI               &   \cite{deb2021}          \\
WC\_MGDA                          & Y          & Gradient                  & HVI               &  \cite{pmlr-v162-momma22a}       \\ \midrule
RLW                               & N          & Scalarization             & $\Delta_m$        &           \\
Uncertainty weighting             & N          & Scalarization             & $\Delta_m$        &  \cite{kendall2018multi}          \\
DynamicWeightAverage              & N          & Scalarization             & $\Delta_m$        &  \cite{liu2019end}         \\ \midrule
MGDA                              & N          & Gradient                  & $\Delta_m$        &   \cite{NEURIPS2018_432aca3a}        \\
GradDrop                          & N          & Gradient                  & $\Delta_m$        &   \cite{chen2020just}        \\
PCGrad                            & N          & Gradient                  & $\Delta_m$        &  \cite{yu2020gradient}         \\
LOG\_MGDA                         & N          & Gradient                  & $\Delta_m$        &   \cite{NEURIPS2018_432aca3a}         \\
CAGrad                            & N          & Gradient                  & $\Delta_m$        &    \cite{liu2021conflict}       \\
LOG\_CAGrad                       & N          & Gradient                  & $\Delta_m$        &    \cite{liu2021conflict}        \\
IMTL                             & N          & Gradient                  & $\Delta_m$        &   \cite{liutowards}        \\
LOG\_IMTL                        & N          & Gradient                  & $\Delta_m$        &   \cite{liutowards}        \\
NashMTL                           & N          & Gradient                  & $\Delta_m$        &  \cite{navon2022multi}          \\
FAMO                              & N          & Gradient                  & $\Delta_m$        &  \cite{liu2023famo}         \\
SDMGrad                           & N          & Gradient                  & $\Delta_m$        &  \cite{xiao2023direction}         \\ \midrule
EC                                & Y          & Scalarization             & HVI               &   \cite{momma}          \\ \bottomrule
\end{tabular}
}
\end{table*}

\begin{figure}[h]
    \centering
    \includegraphics[width=0.95\linewidth]{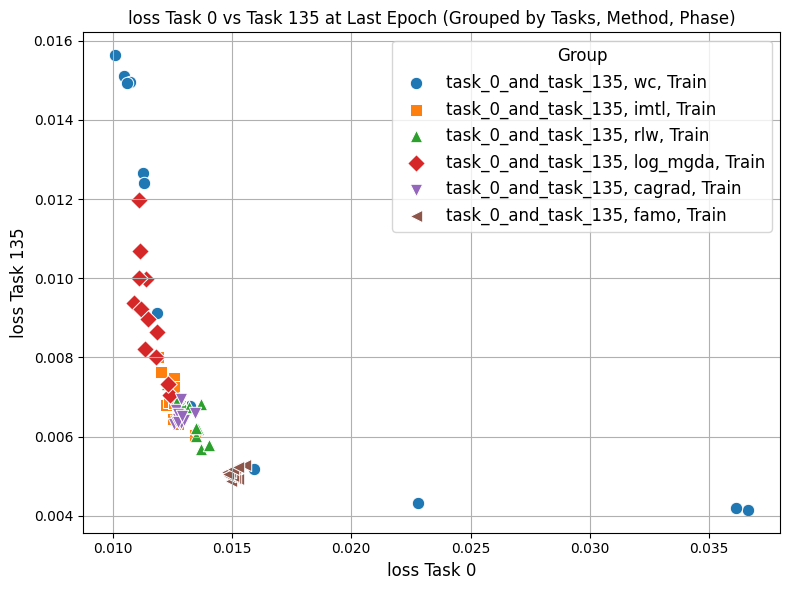}
    \caption{Training losses for 6 selected MTL algorithms.}
    \label{fig:loss}
\end{figure}

\begin{figure}[h]
    \centering
    \includegraphics[width=0.95\linewidth]{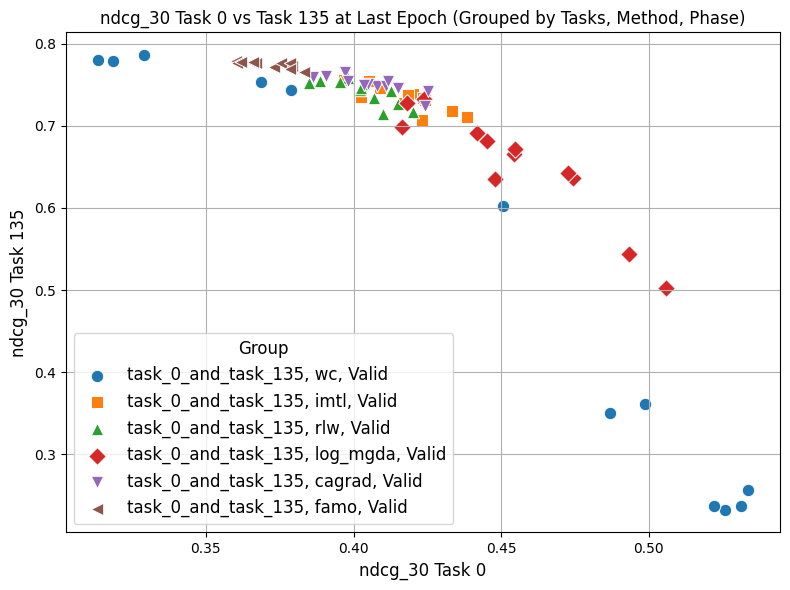}
    \caption{Validation NDCG@30 for 6 MTL algorithms.}
    \label{fig:ndcg}
\end{figure}

\begin{figure}[h]
    \centering
    \includegraphics[width=0.9\linewidth]{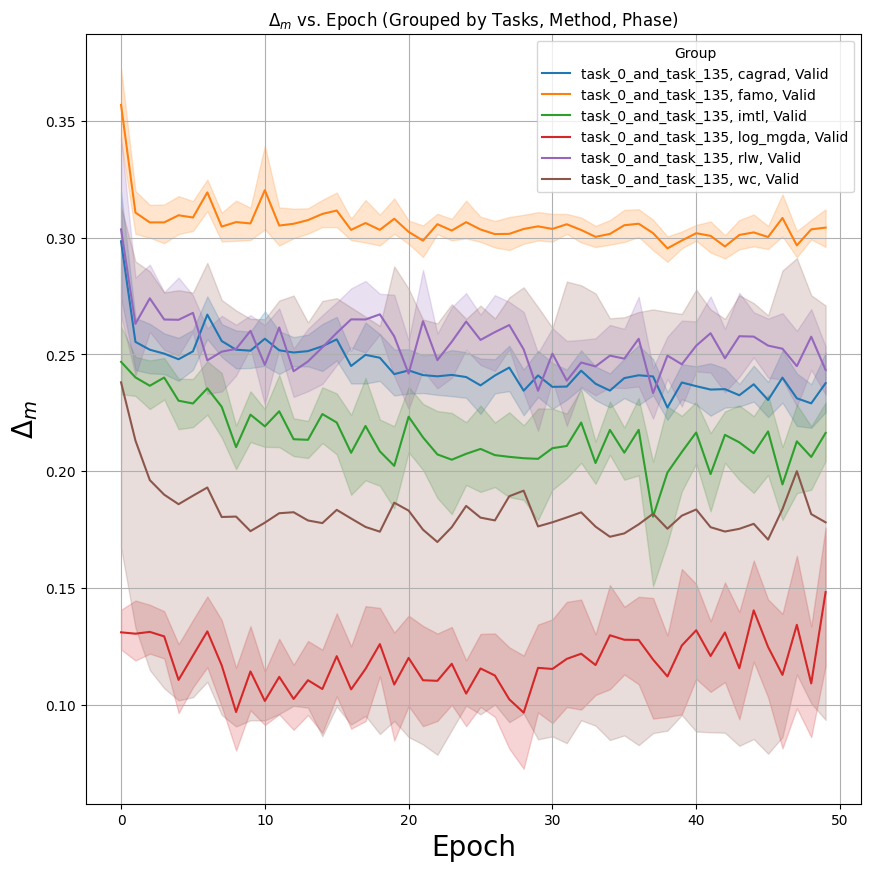}
    \caption{Evolution of $\Delta_m$ for NDCG@30 during validation for 6 MTL algorithms in 10 runs.}
    \label{fig:delta_m}
\end{figure}

\section{Experiment}

\subsection{Dataset and Experimental Settings}

We evaluated DeepMTL2R on the MSLR-WEB30K dataset~\citep{DBLP}. To construct a richer multi-label setting, we followed the protocol of~\citep{momma} and used four of the 136 features as auxiliary relevance labels in addition to the original relevance label (Task~0): Query--URL Click Count (\textit{Click}, Task~131), URL Dwell Time (\textit{Dwell}, Task~132), Quality Score (\textit{QS}, Task~133), and Quality Score2 (\textit{QS2}, Task~135). During training, these feature-derived labels were removed from the model input to prevent target leakage. For the bi-objective setting, we considered all $\binom{5}{2}=10$ label pairs (five labels in total) and, for each pair, ran experiments with 10 different reference vectors.

\subsection{Qualitative Evaluation}
To provide a comprehensive qualitative evaluation, we plot the training losses (Figure~\ref{fig:loss}) and validation NDCG@30 (Figure~\ref{fig:ndcg}) for six selected MTL algorithms across 10 runs. We observe that Pareto-front-finding methods (e.g., WC) recover a clearer Pareto front, whereas non-PF methods tend to concentrate around the central trade-off region. In addition, we report per-task performance and use the aggregate metric $\Delta_m\,\%$ to summarize overall multi-task behavior. The metric $\Delta_m\,\%$ measures the average relative performance change of a multi-task model compared to the corresponding single-task baselines. Formally,
$$
\Delta_m\% = \frac{1}{K}\sum_{k=1}^K (-1)^{\delta_k}\frac{M_{m,k}-M_{b,k}}{M_{b,k}} \times 100,
$$
where $M_{m,k}$ and $M_{b,k}$ denote the performance of model $m$ and the single-task baseline $b$ on task $k$, respectively. Here, $\delta_k=1$ if higher values indicate better performance and $\delta_k=0$ otherwise. As shown in Figure~\ref{fig:delta_m}, WC and LOG\_MGDA achieve the best overall performance among these six MTL algorithms.


\section{Conclusions and Future Work}


We introduced DeepMTL2R, a transformer-based framework for Multi-task Learning to Rank that integrates a broad set of multi-task learning algorithms. By combining context-aware transformer modeling with Pareto-based multi-objective optimization, DeepMTL2R addresses key challenges in multi-task ranking, including capturing list-level context, scaling to large datasets, and balancing competing relevance signals. Experiments on the MSLR-WEB30K benchmark demonstrate the effectiveness of the framework and highlight the trade-offs induced by different MTL strategies. Overall, DeepMTL2R provides a practical and extensible foundation for developing and evaluating multi-task ranking models in real-world settings.

Several directions may further strengthen the framework. First, we plan to extend DeepMTL2R with additional ranking losses and evaluation metrics, including settings with fairness constraints and implicit feedback. Second, incorporating parameter-efficient adaptation techniques (e.g., lightweight task conditioning) may improve scalability as the number of tasks grows. Third, we will investigate more robust preference specification and Pareto-front selection, including automated reference-vector generation and uncertainty-aware model selection. Finally, applying DeepMTL2R to additional domains and datasets (beyond Web benchmarks) will help assess generalization and identify domain-specific design choices.

\bibliographystyle{ACM-Reference-Format}
\bibliography{reference}
\clearpage


\end{document}